\documentclass[12pt]{drpaper}
\title{The New Associationism:\\ Lessons from Deep Learning}
\author{Daniel Rothschild}

\newcommand\blfootnote[1]{%
  \begingroup
  \renewcommand\thefootnote{}\footnote{#1}%
  \addtocounter{footnote}{-1}%
  \endgroup
}
\usepackage{changepage}

\begin{document}

\maketitle
\blfootnote{I am grateful to Emmanuel Chemla, Celia Heyes, Henry Schiller, and  Nick Shea  for extensive feedback. Thanks also to Adrian Alsmith, Emma Borg, David Papineau for discussion.}

\begin{abstract}
What can the success of modern AI tell us about how humans learn? This paper argues that taking AI seriously as a model of human learning supports a modest but genuine associationism. The central finding is that supervised learning---learning driven by evaluative feedback---underlies a surprisingly wide range of contemporary AI systems, from large language models to game-playing agents, differing primarily in how much work is required to generate the relevant feedback signal. This vindicates associationist ideals of a uniform, gradual, error-driven learning mechanism operating across domains, and defuses the once-influential argument that associationist mechanisms are too limited to account for human cognitive capacities. At the same time, the successes of deep learning depend on computational architectures that go well beyond anything classical associationists envisaged, and supervised learning operates within these as one component rather than a complete account of learning.
\end{abstract}

\section{What does AI suggest about the mind?}

AI works. Computers are now performing tasks that, until a few years ago, only humans could do.  Progress in computer science is nothing new. Recognizably modern computers were first built in the mid-twentieth century to perform mathematical calculations far faster and more accurately than humans could. 
Their capacities have been steadily increasing since then. Yet the last few years have seen a sea change in the \emph{kind}  of tasks computers can perform, with striking advances in strategic game play, visual object identification, image generation, voice recognition, and the numerous text-based capacities of contemporary large language models (LLMs).  

My aim here is to examine what recent AI implies for cognitive science. What can we learn from the developments in computer science about how our biological minds work? In this paper, I will focus on human \emph{learning}, asking what AI tells us about how we learn.

As its name suggests, learning is at the heart
of deep learning. Today's deep neural networks begin life with virtually no information about the tasks they will be trained to do. When first initialized they cannot do anything useful.   The appropriate parameter settings, and the performance these enable, must be acquired through extensive training.  It is natural, then, to see today's AI as offering a model of how human learning might work. Like AI systems, humans are prodigies of learning, transforming our minds through extensive training throughout childhood and into adulthood.
That a large swath of our cognitive abilities is acquired through learning is uncontroversial today.\footnote{Even so-called ``nativist'' psychologists like \citet{spelke2022babies} and \citet{Carey2009-CARTOO-3} place learning at the center of human cognitive development.}
It is not the centrality of learning that is at issue in cognitive science today, but its character.

Before turning to \emph{what} we can learn from AI about human learning, let me briefly address the question of \emph{why} we might expect to learn anything at all. Contemporary AI systems are profoundly different from human minds in their training and functioning. They are trained on vastly more data than any human encounters, they use learning algorithms that have no known biological implementation, and they fail in ways that humans do not. These disanalogies are real and cannot be discounted. \footnote{While the resemblance between biological and artificial neural networks is limited, there are nonetheless important basic common traits such as extreme parallelization of processing along with higher-level structural connections between particular network structures and brain structures, like those between convolutional neural networks and mammalian visual cortex. To put it simply: neural networks are more directly inspired by the human brain  than classical symbolic AI systems are.}

The case for taking AI as a model of human learning does not rest on denying these disanalogies, but on a more modest point. We currently lack a well-developed theory of the mechanisms underlying human learning. We know that humans and animals learn, and we know a fair amount about what they learn, but the question of what kinds of computational mechanisms underlie the acquisition of language, object recognition, strategic reasoning, and general inference remains substantially open.  The learning task facing AI systems and the one facing humans are recognizably similar: machines learn to use language, play chess, recognize novel objects. The systems we build to solve problems similar to those humans face in learning are a natural provisional model for our own.

This paper will not directly address the question of how good a model AI systems are for the human brain. Rather it will address the conditional question: \emph{if today's AI systems prove to be useful models for human learning, what do we learn from them?} Whether they are is an empirical matter, to be settled by evidence from developmental psychology, neuroscience, and comparative cognition; at the end of section~\ref{applicationstocognition} I indicate some places where such evidence already points in suggestive directions. The main aim of this paper, however, is the conceptual one: identifying the features of AI learning that the biological story would need to share for the analogy to be informative.

In this paper I will argue for three main claims:

First, I argue that supervised learning is the central learning paradigm in modern AI. Supervised learning, in its paradigm form, is learning from labelled data: the training set consists of input-output pairs, and the system is trained to predict the correct output for new inputs. What the last decade of AI reveals is that this apparently narrow form of learning underlies a surprisingly wide range of AI systems, including many that do not on the surface resemble it at all. These systems differ from the paradigm case primarily in how much intermediate work is required to generate the evaluative feedback that drives learning, but the core mechanism is recognisably the same.

Second, I show how the headline achievements of deep learning, such as novel image generation, language use, and superhuman game-play, are all powered by supervised learning. In each case, what made the breakthrough possible is the reformulation of a wildly different problem so that training data could provide the graded, corrective feedback needed to drive learning. This, I want to suggest, is an underappreciated part of the AI story. Popular accounts tend to credit the recent successes in AI to faster hardware, larger datasets, and more compute. These matter, of course. But they would have counted for little without the prior conceptual work of recognising that language modelling, image generation, and game-playing could all be approached in the same fundamental way. Finding that a single learning mechanism could be made to work across these diverse domains  is no trivial engineering achievement, but rather a conceptual leap forward.

Third, I argue that this pattern of success in AI provides support for a modest but genuine form of associationism. Associationist theories hold that learning proceeds by forming and adjusting associations in response to experience; the classical and operant conditioning paradigms from psychology can be seen as early attempts to model such mechanisms. For much of the twentieth century, associationism was on the retreat, outpaced by the apparent success of rule-based and symbolic approaches to cognition. That the deep learning story provides some vindication of associationist ideals has been widely noted. What I aim to do is say more precisely what that vindication amounts to, and where it runs out. The gradual, error-driven adjustment that associationists pointed to turns out, when implemented at scale, to be extraordinarily powerful. But the successes of deep learning also depend on computational architectures that go well beyond anything associationists envisaged, and supervised learning operates within these architectures as one component rather than as a complete account of learning. The result is a picture that is associationist about the core learning mechanism while acknowledging that the mechanism alone does not explain the full story.

Before moving to these claims, in the next section I  discuss the success of connectionist models and the question of what level of abstraction it is most fruitful to pitch the lessons we draw from AI.

\section{Connectionism and levels of explanation}

One obvious lesson from recent AI is that the promise of the computational paradigm variously called \emph{connectionism}, \emph{parallel distributed processing} (PDP), or \emph{artificial neural networks} (ANNs) 
has finally been realized by the accomplishments of its newest incarnation, \emph{deep learning}. The value of the connectionist programme, as a model of human cognition and learning and as a useful computational tool in its own right, was fiercely debated among cognitive scientists from the late 20\textsuperscript{th} to the early 21\textsuperscript{st} centuries. The cognitive scientists who developed it viewed it as a game-changing model for much of human cognition. As early as the mid-1980s, the main proponents of connectionism wrote that neural networks
``hold out the hope of offering computationally sufficient and psychologically accurate mechanistic accounts of the phenomena of human cognition which have eluded successful explication in conventional computational formalisms; and they have radically altered the way we think about the time-course of processing, the nature of representation, and the mechanisms of learning'' {\citet{appealparallel}}.

Skepticism about  connectionism, though, was widespread among philosophers, psychologists and computer scientists.\footnote{\citet{fodor1988connectionism} and \citet{pinker1988language} are classic criticisms of connectionism.} 
A quarter of a century ago 
it was possible to dismiss connectionism 
with remarks like this:
\begin{quote}
The successes of the distributed connectionist program have been limited \ldots mostly being confined to various forms of pattern recognition; and there are principled reasons for thinking that such models cannot explain the kinds of structured thinking and one-shot learning of which humans and other animals are manifestly capable. \citep{Carruthers2002-CARTCF-2}
\end{quote}
In an era in which connectionist systems beat humans at chess and Go, solve Math Olympiad tests, and write poetry, the 
opponents of connectionism seemed to have backed the wrong horse.  

That the connectionist programme has succeeded seems obvious. My goal  is to home in on the specifics of what we might learn about how  humans learn from the success of current connectionist models. That is, which features of learning in today's connectionist systems might we plausibly think also obtain in biological minds? In other words, what more nuanced take-home message can we find beyond the generic “Yay, connectionism!”?

One natural way to spell out the lesson from the success of connectionism is to take the structure of contemporary neural networks as  models for human cognition. 
 Neural networks come in many shapes and sizes and are trained with different regimes. The term \emph{deep learning} refers to a particular type of network architecture and training regime \citep{deeplearninglecun}. The successes of deep learning are not the successes of ANNs generally, but of many-layered artificial neural networks trained with backpropagation, 
typically in specific forms like convolutional networks, deep Q-networks, and transformers. So the successes in AI naturally would support treating such connectionist systems as models for human cognition. 

There is nothing wrong with such a mechanistically specific lesson. The distinctive approach of connectionism is often phrased at this level: connectionism is a model of a particular kind of computation. As a representative example, in his position paper \citet{bechtel1988connectionism} describes connectionism as follows:
\begin{quote}
The basic components of the connectionist architecture are simple \emph{units} which, like neurons, are, at any given time, \emph{activated} to some degree. Typically, this activation consists in possessing an electrical charge. These units, again like neurons, are \emph{connected} (these connections can be of varying
strengths) to other units so that, depending on their own activations, they can act to increase (excite) or decrease (inhibit) the activations of these
other units. Additionally, in some connectionist systems, these connection strengths can be altered as a result of activity in the system so that the effect
of one unit on another can change over time.
\end{quote} This passage reveals a perspective in which the key to connectionism is the detailed structure of the computational system.

There are, however, downsides to a narrow focus on the mechanistic level.\footnote{``Mechanistic'' is used here in a narrow, implementational sense---the specific computational architecture (units, activations, connections) and training algorithm. Some associationists use ``mechanism'' more broadly, to cover everything internal that contributes to behaviour; on that usage, the functional features I distinguish would also count as mechanistic.} Describing the structure of networks in terms of units, activations, and weighted connections tells us relatively little about what, \emph{functionally speaking}, these systems are doing as learners. If instead we focus on the \emph{learning problems} the systems solve, and on the abstract form of the learning procedures they use, we may obtain a different and in some respects more informative picture of what is going on in contemporary AI. Moreover, by looking at AI at this functional level, we can draw analogies between human cognition and AI that are not hostage to implementational details and thus might be more plausible as claims about biological minds than mechanistically specific analogies would be.

In the rest of the paper I pursue this higher-level perspective.\footnote{In Marr's terms \citeyearpar{MarrVision}, this is the \emph{computational} level of analysis. The claim is about subpersonal computational processes---the kinds of operations that might underwrite cognitive capacities---rather than about anything accessible at the level of conscious thought.} Rather than asking whether human brains implement something like a transformer or a convolutional network or indeed any kind of deep neural network, I focus on the general structure of learning in today's AI. 

\section{The nature of learning in AI}\label{naturelearning}

The successes of deep learning vindicate what can be called the \emph{statistical} or \emph{machine-learning} approach to learning.\footnote{\citet{bishop2006pattern} is a classic textbook for machine learning.}
Machine learning, generally, concerns itself with finding and exploiting statistical patterns in data. Machine learning is a broad area of computer science with many methods, and there is no precise definition, nor any universally accepted way to contrast it with other forms of learning. To claim that machine learning is the dominant approach to learning in AI is nonetheless to make a substantive point. We can think of learning, in general, as any computational process that uses input data to improve performance.\footnote{See, for example, \citet{Weiskopf2008-WEITOO} and \citet{Margolis2011-MARLMT-2} for refinements of this general definition of learning.} In this sense, plenty of learning is unrelated to machine learning. A classical AI program designed to play chess is learning in the general sense when it sees its opponent's move and chooses its reply. Even a word processor that changes the font when the user selects a new one is learning. But these are not examples of statistical learning in any meaningful sense; they are pre-programmed responses to specific input. Machine learning techniques are irrelevant in these cases. Machine learning, by contrast, involves inference from a substantial pool of data, with a large and generally continuously varying range of responses to the data available. It is this statistical style of learning that now dominates AI. 

The fact that connectionist systems work only by capturing statistical patterns was often cited as a weakness in the classical debates of the late 20\textsuperscript{th} century. 
When \citet{fodor1988connectionism} wrote that ``Connectionism treats learning as basically a sort of statistical modeling,'' they pointed to what they saw as a major limitation of the programme.
Fodor and Pylyshyn argued that such ``frequency-sensitivity'' would fail to acquire the systematic rules needed for much human cognition. 
While there is room for debate on how well deep learning systems capture the systematic and compositional aspects of human thought, the accomplishments of LLMs and other systems suggest that the once-derided connectionist ``statistical modelling'' is, in fact, very powerful.

We would do well to be more precise about what kind of statistical reasoning underpins contemporary connectionism (i.e.\ deep learning). 
After all, many approaches to learning are statistical in some sense. 
\citet{fodor1988connectionism} contrast connectionism with what they regard as the traditional account of learning, \emph{hypothesis testing}: 
\begin{quote}
There is an alternative to the Empiricist idea that all learning consists of a kind of statistical
inference, realized by adjusting parameters; it's the Rationalist idea that some learning is a kind
of theory construction, effected by framing hypotheses and evaluating them against evidence.
\end{quote}
However, even their ``rationalist'' approach retains a statistical element: evaluating hypotheses against evidence. Putting aside Fodor's rabidly nativist approach, there are many statistical approaches to learning, such as the Bayesian programme in cognitive science \citep[e.g.][]{tenenbaum2011grow} that are very far in spirit from connectionism. In practice Bayesian learning models require specification of a complete hypothesis space, along with a set of priors over them. Bayesians working in cognitive science tend to specify the hypothesis space using traditional, symbolic AI.  Connectionist models, by contrast, do not require explicit specification of a hypothesis space.\footnote{There is more to be said here. It is possible to give a Bayesian interpretation to the learning of neural networks. My only point here is just that statistically grounded accounts of learning can have more of the flavour of classical, symbolic AI than connectionist approaches do.} 

What I will do in the next section is delineate the particular form of statistical learning found in deep learning. 

\section{Supervised Learning}\label{supervised}

Most introductions to machine learning begin with \emph{supervised learning}, a type of learning in which the training set consists of \emph{labelled data}: pairs of an input (an example or instance) and an output (a label). It is easiest to understand through an example. A simple illustration is a collection of images, each labelled ``cat'' or ``non-cat''. The aim of the learning system is to find a process that makes accurate predictions on new, unseen images---not just the ones in the training set. When the labels take discrete values the task is called \emph{classification}; when they vary continuously it is called \emph{regression}. Training proceeds by scoring the system's outputs against the correct labels and adjusting the system's internal parameters to reduce the gap---a process described in detail in section~\ref{deeplearning}.

Supervised learning can seem a rather narrow type of learning. When we consider how humans and other animals learn, most of their learning does not resemble it. We do not master mathematics simply by generalising from large sets of question--answer pairs; we are taught underlying rules and principles. Many skills---say, writing a poem---are acquired merely by observation, with no explicit input--output labels attached to the examples. Richard Sutton, a pioneer of reinforcement learning, went so far as to claim that ``Supervised learning is not something that happens in nature.''\footnote{Dwarkesh podcast, 26 September 2025. The full quote: \begin{quote} Supervised learning is not something that happens in nature. Even if that were the case with school, we should forget about it because that's some special thing that happens in people. It doesn't happen broadly in nature. Squirrels don't go to school. Squirrels can learn all about the world. It's absolutely obvious, I would say, that supervised learning doesn't happen in animals.\end{quote}}

It is not surprising, then, that machine learning includes several other paradigms besides supervised learning. There is a diverse collection of methods called \emph{unsupervised learning}, of which clustering is a classic example. A third major paradigm is \emph{reinforcement learning}, in which a system learns by choosing actions that maximise a cumulative reward signal over time.

Despite being one of many paradigms, supervised learning plays an outsized role in contemporary machine learning.\footnote{Its dominance can be seen by how much of major textbooks like \citet{hastie2009elements} and \citet{bishop2006pattern} are devoted to supervised learning.} One part of this paper's aim is to explain why a seemingly limited form of learning has proved so dominant, and what this tells us about human cognition.

The key to understanding the reach of supervised learning is to notice that what the labelled data provides is a way of evaluating how well the system is currently performing and thereby guiding improvement. The labels themselves need not come pre-packaged by a human annotator. What matters is that training data of some kind supplies the targets needed to evaluate the system's current guesses and correct them. This observation points to a family of learning problems that share the essential structure of supervised learning while differing in how the labels are obtained. The paradigm case is images paired with human-assigned labels. At varying distances from it lie methods where the labels are generated from the data itself, by processing of increasing complexity. At the near end of this spectrum, the labels are extracted from the raw data by a trivial transformation---no human annotation required. Further along, generating the supervision requires considerably more theoretical ingenuity: the required targets are not derivable from the training data directly, but must be constructed through intermediate computations that are themselves part of the learning process. The next section examines the most important cases across this spectrum in detail.

It is worth noting, however, that the mere possibility of reformulating a problem as a supervised one is a weak and uninformative claim. In a trivial sense almost any learning problem can be given some evaluative signal; what matters is whether that signal can be made precise and tractable enough to drive learning effectively. The significant finding about contemporary AI is not that these problems can be cast as supervised learning in principle, but that so many of them can be cast in a form that is feasible in practice, yielding systems that actually learn. Sutton was right that supervised learning, in its most literal form, does not occur in nature. What the last decade of AI suggests is that nature may nonetheless have found ways to cast the actual problems it encounters as supervised learning problems---a suggestion I return to at the end of section~\ref{applicationstocognition}.

\section{Deep Learning as the Engine of Supervised Learning}\label{deeplearning}

The term ``supervised learning'' is standardly used in AI both for a class of learning problems---those involving labelled data---and for the family of methods used to solve them. Identifying a learning problem as a supervised one, or as reducible to a supervised subproblem, does not by itself tell us how to solve it. A separate question is what computational machinery is used to find a function that performs well on the training data and generalises to new inputs. This section describes the dominant answer in contemporary AI, and why it matters.

The standard method for solving supervised learning problems today is \emph{gradient descent}. The idea is to represent the target function using a large set of adjustable numerical parameters, start with some initial setting of those parameters, and iteratively improve them by computing, at each step, which direction of adjustment reduces the system's current errors on the training data. This requires a \emph{loss function}: a measure of how poorly the current parameter settings perform, computed from the discrepancy between the system's outputs and the correct labels. Gradient descent follows the direction in which small adjustments to the parameters most reduce this loss, making incremental improvements over many iterations.\footnote{See \citet{bishop2006pattern} for a thorough treatment.} The process is the same whether the labels are human-assigned, automatically generated, or bootstrapped from intermediate predictions: in each case, learning consists in adjusting parameters to reduce a loss.

What is distinctive about gradient-based learning is that the evaluative signal is local and per-example: the system is scored on how well its current parameters perform on individual input-output pairs, and adjustments can be made incrementally in response. This differs from search paradigms that operate over explicit hypothesis spaces, such as Bayesian inference over probability distributions or classical AI search over symbolic rule systems, where the evaluative signal operates at the level of whole hypotheses rather than individual instances. That so many cognitive tasks admit of this per-example treatment is not obvious in advance. It is a substantive finding about the structure of those problems: that they can be framed as finding a good input-output mapping, and that local scoring provides sufficient information to guide that search effectively. Supervised learning is the paradigm case where this scoring is most direct, but the same architecture underlies the wider family of methods discussed in this paper, including those where the connection to supervised learning is more indirect.

Gradient descent is not specific to deep learning. Linear regression, kernel methods, and many other techniques optimise a loss over training data. What distinguishes contemporary deep learning is the class of functions used to represent the learned mapping: \emph{deep neural networks}, networks with many successive layers of simple computational units whose connection weights are the parameters being optimised. Training is carried out by \emph{backpropagation}, an efficient algorithm for computing the gradient of the loss with respect to all parameters simultaneously, even in very large networks.

Two properties of deep networks are particularly relevant to their dominance. First, they are highly expressive: sufficiently large networks can approximate a very wide range of input-output mappings, far beyond what shallower or more constrained models can represent. Second, deep networks are organised into successive layers, and there is evidence that in at least some domains---particularly vision---these layers come to encode increasingly abstract features of the input, from edges and textures to object parts. Whether this generalises across all the domains in which deep networks succeed is less clear, but this hierarchical organisation appears to be part of what enables effective generalisation.

Deep neural networks are the machinery that solves these supervised learning problems. The paradigm cases from the previous section, such as image classification, language modelling, value prediction in reinforcement learning, are all solved by deep learning. What the last decade has revealed is how many problems, once suitably reformulated as supervised learning problems, can be handled by this single engine. The next section examines the most important cases in detail.

\section{Supervised Learning in Action}\label{applicationstocognition}

The story of the success of deep learning is often told as a triumph of scale: faster hardware, larger datasets, more compute. These are certainly key ingredients. But equally important, and less often remarked on, is the theoretical ingenuity by which a diverse range of learning problems have been reformulated so that supervised learning can be brought to bear on them. What follows examines  some of the most significant cases, arranged roughly by the distance between the original problem and the supervised learning subproblem at its core.

Language modelling sits close to the supervised end of the spectrum, but the intellectual move that made it work was far from obvious in advance. The challenge of acquiring language from text might seem to call for rich structured supervision: grammatical rules, semantic annotations, explicit feedback on errors. What the developers of large language models discovered instead was that a deceptively simple prediction task suffices. A transformer network is trained to predict each masked token from its surrounding context, with the correct token serving as the label. This is supervised learning in the straightforward sense, with labels generated automatically from the raw text. The surprise is not the mechanism but the result: that training on next-token prediction at scale produces systems capable of translation, summarisation, reasoning-like behaviour, and general-purpose conversation. The cognitive range of what emerges from such austere supervision is one of the more startling discoveries in deep learning. 

Image generation presents a different challenge. Recognising whether an image contains a cat is one thing; generating a realistic image of a cat from scratch is another. The dominant approach today, \emph{diffusion models}, achieves this through a similarly elegant reformulation \citep{sohl2015deep}. Training data is corrupted by adding known amounts of random noise, and the network is trained to predict the noise given the corrupted image. Again the labels are generated automatically, by construction. A network trained in this way can then generate novel images by starting from pure noise and iteratively removing it. The generative capacity emerges entirely from supervised training on a denoising task.\footnote{In leading implementations such as Stable Diffusion, the diffusion process operates not in pixel space but in a compressed latent space learned by a variational autoencoder.}

The most striking reformulation is in reinforcement learning. Consider learning to play chess at superhuman level. The raw learning signal here is maximally sparse and non-directional: win or lose, after a long sequence of moves, with no indication of which moves were good or bad. There is no obvious path from this signal to a tractable supervised learning problem. The theoretical contribution of temporal difference learning is to construct one. Rather than treating the problem as one of finding good moves directly, TD methods train the system to predict future value: how promising is the current board position likely to prove? This prediction problem is supervised in the relevant sense, with targets bootstrapped from the system's own value estimates at later time-steps \citep{sutton1998reinforcement}. Once value prediction is learned well, it can guide action selection. The ingenuity lies in seeing that the problem of learning to act well can be approached via the problem of learning to predict well.\footnote{\citet{tesauro1994td} developed the first network-based reinforcement learning system along these lines, for backgammon. DeepMind extended the approach to video games, chess, and Go \citep{DBLP:journals/corr/MnihKSGAWR13,10.1038/nature24270}.}

A clarification is worth reiterating from section \ref{supervised}. It is trivial to observe that any learning problem can, in principle, be given some evaluative signal, and thus framed as a form of supervised learning. The situation I've outlined is non-trivial: the specific problems identified here admit reformulations that are not merely formally supervised but tractably solvable as supervised learning problems by current systems.\footnote{Indeed, even being tractable by formulating as a problem in which family of parametrized functions is optimized by gradient descent is a non-trivial property. Gradient descent is not the only way to learn.} Not all problems do. Even with extensive training data, deep networks today do not recover Newton's laws from observations of physical systems obeying them. The breakthroughs in language modelling, image generation, and game-playing were breakthroughs precisely because the reformulations found were ones that deep networks could actually solve.

It is worth closing this section by asking whether these reformulations have biological counterparts. Sutton's claim, recall, was that supervised learning does not happen in nature. This claim deserves to be taken seriously, and in its most direct form it is probably correct: nature does not supply organisms with pre-labelled training sets. But my argument has been that supervised learning operates not only in its direct form, but as a component within more complex learning processes. The question then becomes whether those more complex processes have biological counterparts---and here there is suggestive evidence. The predictive processing tradition holds that the brain continuously generates predictions of its sensory input and updates itself on the basis of prediction errors \citep{clark2013whatever, hohwy2013predictive}, which has the structure of a self-supervised problem: the next moment of experience provides the label for the prior prediction. And Sutton himself, with Barto and others, has documented that dopaminergic signalling in the mammalian brain tracks reward prediction error in a way that closely parallels temporal-difference learning \citep{schultz1997neural, niv2009reinforcement}---the very mechanism by which supervised learning is embedded within reinforcement learning in AI. The point is not that these proposals are uncontroversial, but that the strategy of constructing supervised subproblems from experience, rather than receiving labels directly, may be exactly what biological learning does. Sutton may be right that pure supervised learning does not happen in nature. What the AI story suggests is that something structurally very close to it does.

\section{Associationism in the age of AI}\label{assoc}

Perhaps the most striking lesson from AI today is how a single basic mechanism is able to do so much for the learning process.   The idea of a central mechanism of learning has a long history in philosophy and psychology, going back at least to the British empiricists and moving on to the vibrant associationist tradition in psychology which reached its heights in the mid-twentieth century before the cognitive revolution. The links between neural networks and associationism, in its various forms, are long and deep.\footnote{See, for example, \citet[ch. 4]{shanks1995psychology}}

Associationism is a family of views in philosophy and psychology according to which the brain learns by forming associations in response to experience. 
Like connectionism itself, associationism is often understood at a mechanistic level. For example, \citet{heyes2018cognitive} writes, ``Associative learning can be very broadly defined as learning in which
an excitatory or inhibitory link is formed between representations of
events.'' Many different kinds of  learning systems might be implemented by systems  strengthening and weakening links between representations. What we want, in the first instance, is to view associationism in fairly abstract, functional terms.

The easiest way to get this perspective on associationist learning is to look at the two major forms of associationist learning in psychology: \emph{classical} (or \emph{Pavlovian}) \emph{conditioning} and \emph{operant conditioning}. In classical conditioning, statistical regularities in inputs are mimicked by statistical regularities in internal states.  Pavlov's dog learns to transition from a representation of the bell to a representation of the food that reliably follows it. Classical conditioning  is fundamentally about prediction: creating associations that allow internal states to mimic regularities in the world.  In operant conditioning, by contrast, \emph{actions} are shaped by their consequences: behaviours that lead to greater reward become more frequent. This form of learning relates more closely to the machine learning paradigm of reinforcement learning. However, operant conditioning also depends on learning environmental contingencies, and  is thus not entirely distinct from classical conditioning.\footnote{Indeed the relationship between the two types of associationist learning was a long-standing topic in the associationist literature \citep[see, e.g.,][]{rescorla1967two}.} 

As we saw in section \ref{applicationstocognition}, supervised learning can be used to power both forms of associative learning. Classical conditioning,  understood as a problem of prediction, can be addressed by self-supervised learning such as that used in LLMs. More generally, one can think of supervised learning as a process of ``associating'' inputs with outputs through learning.  As long as the system receives directed feedback in response to its outputs, it can gradually improve by gradient descent. Supervised learning---as implemented by gradient descent---is a type of learning well suited to implement classical associationism.

With respect to operant conditioning, the situation is somewhat more complex. Operant conditioning is about the control of behavior, not merely prediction. It is learning how to act so as to increase reward. Its closest kin that we have discussed is not supervised learning itself but rather reinforcement learning. Indeed, the connection between reinforcement learning and associationism has long been recognized, with aspects of reinforcement learning being inspired by associationist psychology \cite[see][particularly chapter 14]{sutton1998reinforcement}. As \citet{sep-associationist-thought} write , ``RL can be seen as formalizing the core problem of associationist theories of learning: how an agent learns to select beneficial actions by associating stimuli and responses based on their experienced consequences.'' As we saw in section \ref{applicationstocognition}, in practice, reinforcement learning in deep learning today is accomplished via mechanisms that include supervised learning. 

So supervised learning is intimately connected to both major forms of associationist learning. The success of  supervised learning is naturally thought of as a result that supports the overall associationist programme, if not its detailed implementation. 

This vindication of associationism goes beyond the mere fact that associationist learning (of both main sorts) is closely linked to the forms of learning in deep learning. Associationism is not just committed to these types of learning being central to human learning, but also to a perspective on how this learning is implemented in the humans and animals. There are at least two central themes in associationist thought, that we can think of as vindicated by the particular way in which supervised learning has become central to AI today:
\begin{description}

  \item[Uniformity]\hfill\\
The \emph{uniformity} of learning mechanisms across different learning tasks counts as a vindication of the associationist paradigm. A central tenet for many associationists is that similar basic mechanisms underlie learning in different contexts. Associationists posit that one or two core learning processes suffice to explain a wide variety of learned capacities. This claim of a uniform learning mechanism runs through early associationists such as Thorndike and Skinner, and into contemporary cognitive associationism.\footnote{Thorndike held that his laws of learning ``stand out clearly in every series of experiments on animal learning and in the entire history of the management of human affairs'' \citep{thorndike1911animal}. Skinner extended operant conditioning to the full range of human behavior---including language, social behavior, and culture \citep{skinner1953science}. While the contemporary cognitive associationist \citet{shanks1995psychology} writes ``\ldots associative learning is at the heart of any organism's psychological capacities.''}

It is worth being clear about the level at which this uniformity operates. Many accounts of learning are uniform at some level of abstraction: Bayesian accounts take all learning to proceed by conditionalization, and hypothesis-testing accounts take it to proceed by framing and evaluating hypotheses. But in each case the uniformity is at a high level of abstraction---the concrete learning procedures may differ radically across domains, since each requires its own hypothesis space, priors, or candidate hypotheses. The associationist claim is stronger: that the core learning rule is itself the same across domains, not merely instantiated differently within a common schema.

Deep learning can be seen as a vindication of this stronger claim: the same basic supervised learning mechanism---namely the gradient-based training of deep networks---is deployed across vastly different domains, including vision, language, game play, and more. This functional generality offers a kind of proof of concept for the generalist, associationist vision.

\item[Gradual, continuous, error-directed learning]\hfill\\
  Associationist theories typically understand learning as a gradual, statistical process in which predictions improve with experience. This tends to be captured mechanistically by discussing small changes in associations in response to experience. Such small adjustments are also a defining feature of  supervised learning via gradient descent as well. Gradient descent is a form of learning-by-error-correction: small adjustments are made iteratively to reduce prediction error, just as is posited by influential accounts of classical conditioning \citep{RescorlaWagner}.\footnote{Rescorla himself drew this connection explicitly: ``Connectionistic theories of this sort bear an obvious resemblance to theories of Pavlovian conditioning. Both view the organism as using multiple associations to build an overall representation, and both view the organism as adjusting its representation to bring it into line with the world, striving to reduce any discrepancies\ldots the so-called delta rule, is virtually identical to one popular theory in Pavlovian conditioning, the Rescorla-Wagner model. Both are error-correction rules, in which the animal uses evidence from all available stimuli and adjusts the strength of each stimulus based on the total error'' \citep[p.~159]{RescorlaPavlovian}.}

\end{description}

Importantly, there are also features defined by what associationism excludes that find support in deep learning. Associationist learning is  characterized by \emph{not} having classic symbolic systems facilitating learning. The training of neural networks by backpropagation also does not look like a process of a computational system making symbolic inferences.\footnote{Of course, there may be a higher level of abstraction on which the training of a specific network  can be understood in more classical terms. Ruling out this kind of story is difficult, if not impossible. But the fact that we can directly program neural networks to learn from data without explicitly building classical structure into the learning routine is at least \emph{evidence} against the idea that the learning process is best understood in more abstract, classical terms.} So the apparent absence of clearly non-associationist mechanisms of learning within supervised learning implemented by deep neural networks can be seen as supporting the associationist programme.

As we have seen, the connections between successes of supervised learning in AI and associationism are strong but not entirely straightforward. It is not the case that, on the surface, deep learning simply is associationist. It is rather a combination of different connections: standard associationist mechanisms (classical and operant conditioning) can be implemented easily with deep learning, in addition, the associationist ideal of positing one continuous and gradual, general mechanism of learning that applies across domains is realized in deep learning by the prevalence of supervised learning via backpropagation in today's deep neural networks.

Perhaps the strongest support for associationism comes from the near-collapse of what had been a major argument against associationist models: that they are simply not powerful enough to account for the kind of learning that humans are capable of. The idea that there are (to repeat the quote from Carruthers) ``principled reasons for thinking that [connectionist models] cannot explain the kinds of structured thinking and one-shot learning of which humans and other animals are manifestly capable''  is no longer plausible. The same class of mechanisms critics conceded might handle low-level pattern recognition turns out, when properly implemented and scaled, to also be capable of handling much more sophisticated feats of reasoning.

\section{Limits of associationist learning}\label{beyondassoc}

While the triumph of supervised learning in AI provides support for associationism, in this closing section, I make three points which add nuance to the story. The first concerns the relationship between training and inference in contemporary AI systems. The second is that the successes of deep learning have been driven by computational architectures which go beyond anything envisaged by mainstream associationists. The third is that the successes of deep learning do not provide much support for the empiricist idea that most human learning proceeds in a domain-general fashion.

A natural worry about the associationist picture I have been developing concerns the behaviour of contemporary AI systems at inference time. When a large language model works through a complex problem step by step, or sustains a long chain of reasoning across many exchanges, it can seem to be doing something quite unlike the gradual, error-driven weight adjustment that characterises associationist learning. Does this not show that the most cognitively impressive capacities of these systems are not associationist at all?

The worry partly rests on a conflation of two distinct phases. There is, on the one hand, the slow learning that occurs during training: the iterative adjustment of parameters by gradient descent over vast quantities of data. This is the process I have been arguing is largely associationist in character. There is, on the other hand, the fast processing that occurs at inference time: the deployment of the trained network to handle new inputs, including the kind of extended in-context reasoning that strikes observers as distinctively non-associationist. It is striking that slow associationist-style training suffices to produce systems capable of fast, flexible, apparently reasoning-like behaviour. What associationist training produces, however, is not itself associationist.\footnote{The gap between what is learned and how it is expressed in behaviour has long been recognised within the associationist tradition itself. Surveying the state of Pavlovian conditioning, \citet[p.~158]{RescorlaPavlovian} concluded that ``conditioning involves the learning of relations among events that are complexly represented, a learning that can be exhibited in various ways. We are badly in need of an adequate theory of performance in Pavlovian conditioning, but the classical notion of a new stimulus taking on the ability to evoke an old response clearly will not do.''} The systems that emerge from gradient descent over vast data engage in something that looks much more like reasoning than like conditioning. Associationism about learning mechanisms is thus compatible with, and indeed helps explain, cognitive outputs that would have seemed to refute associationism entirely.

My second point concerns the existence of structures that go well beyond the associationist framework in deep learning. AI's success is not just about big neural networks being fed lots of data. It is also about very specific computational structures within which these networks are embedded, and which shape what they can learn.\footnote{\citet{SmolenskyMcCoy} emphasize that compositional structures are key to AI successes: \begin{quote}
Thus both of these truly seminal network types---CNNs and Transformers---derive
much of their power from their additional compositional structure \ldots:
spatial structure and a type of graph structure.
\end{quote}}
In convolutional neural networks (CNNs) a particular computational structure is imposed by constraints  on the network. The constrained structure forces the network to process all small areas of the visual input in the same way \citep{lecun1998gradient}. It is only with this kind of structure, modeled on the mammalian brain, that neural networks can feasibly be trained to perform visual recognition.
Likewise, diffusion models, while trained via a supervised denoising objective, operate within an architectural framework that is not itself supervised learning: the iterative noise-removal process at inference time has no counterpart in the training procedure. The successes in AI are the product of sophisticated computational architecture rather than simply scaling up speed and data. These architectures include elements of supervised learning within them, but the whole system is not itself an instance of supervised learning.

To put this in perspective, twentieth-century associationist psychologists usually distinguished only two kinds of associative learning: operant and classical conditioning. But if we were to look for the types of learning from what is happening in AI today, it would be natural to posit many more types of learning. A deep neural network learning to classify via back-propagation corresponds closely to classical conditioning. The deep Q-networks used for reinforcement learning correspond closely to operant conditioning. But the kinds of learning that we see in various paradigms in deep learning such as diffusion models  for example, do not fit into either category.\footnote{Other deep learning models that do not cleanly fit into the associationist paradigms are Generative Adversarial Networks (GANs) and Variational Autoencoders (VAEs).} Like reinforcement learning, they \emph{use} supervised learning, but they have their own distinctive computational profile, with little direct precedent in the rich history of associationist psychology.

In sum,  supervised learning implemented by deep neural networks is a crucial technology used in AI, but it is embedded in computational structures with distinct rules. In the case of LLMs, for example, while the basic problem being solved is a supervised learning one, namely, predicting missing words, the transformer architecture with its attention heads gives considerable structure to the form of learning \citep{transformer}. So a strong form of associationism, in which most learning mechanisms are simple associationist ones, is not vindicated by the advances in deep learning today. Supervised learning, as we saw, only works as \emph{part} of larger, more complex learning mechanisms that go beyond the limited computational vision one can find in the associationist literature.

My third point concerns empiricism. Associationism and empiricism have long been fellow travellers in philosophy and psychology, and it is worth being clear about the relationship between them here. Empiricism, in its strong form, is the claim that knowledge and conceptual content derive from experience rather than being innately specified. Associationism is a claim about learning mechanisms: that the brain acquires capacities by forming and adjusting associations in response to experience. These positions are connected but distinct. One can hold that associationist mechanisms are central to learning while also holding that those mechanisms operate on innately structured input representations, or that they are supplemented by domain-specific inductive biases. The argument of this paper is associationist in the mechanistic sense. It does not straightforwardly entail an empiricist thesis.

The question of domain-generality is where this distinction matters most.\footnote{Here I follow many, including, \citet{Carey2009-CARTOO-3}, \citet{Margolis2013-MARIDO-11} and \citet{Buckner2023-BUCFDL}.} The success of supervised learning across vision, language, and strategic reasoning supports the claim that a common learning mechanism operates across domains. It does not, by itself, show that this mechanism operates without domain-specific structure or domain-specific inductive biases. The evidence from developmental psychology for domain-specific inductive biases in areas like number, agency, and spatial reasoning \citep{DehaeneNumber, CareyDaedalus, spelke2022babies} is not undermined by the success of domain-general learning mechanisms in AI. Shared mechanisms across domains and domain-specific biases are compatible, and the extent of each in biological minds remains an open empirical question.\footnote{Buckner's \citeyearpar{Buckner2023-BUCFDL} rich discussion of AI and empiricism covers related ground. Buckner argues that the success of artificial neural networks supports a form of empiricism, emphasizing in particular that different architectures prominent in AI can be applied across psychological domains. My emphasis has been somewhat different: that these architectures share, under the hood, a common learning mechanism in supervised learning implemented by gradient descent. Both points speak to domain-generality of mechanism. Where I would part from a strong empiricist reading is in treating the domain-generality of mechanism as settling the domain-generality of the overall learning system. The question of how much domain-specific structure and inductive biases biological learners bring to the task remains substantially open. One substantive proposal comes from \citet{HeyesChaterDwyer2020}, who argue that evolved domain-specific adaptations are concentrated in peripheral cognitive processes (perception, attention, motivation, motor control) rather than in central mechanisms of learning and inference. On their view, the central/peripheral distinction long recognised in associationist psychology allows domain-specific cognitive organisation to be compatible with a broadly domain-general central learning mechanism---a structural parallel to the picture developed in section~\ref{beyondassoc}, on which supervised learning by gradient descent sits at the core of diverse computational architectures.} Even if AI successes give limited direct empirical support for empiricism, what they do is undermine the old argument---beginning at least with Plato's \emph{Meno} and influentially echoed by \citet{FodorLOT}---that nativism must be true because there can be no viable account of learning.

The picture that emerges is neither a wholesale vindication of associationism nor a refutation of it. Supervised learning, implemented by gradient descent in deep neural networks, turns out to be the central ingredient in a remarkably wide range of AI systems --- including many that, on the surface, look nothing like it. This is a substantive finding, and it lends genuine support to the associationist vision of a single, general, error-driven learning mechanism operating across domains. At the same time, the successes of AI depend on computational architectures that go well beyond anything associationists envisaged, and the associationist core operates as one component within these larger structures rather than as a complete account of learning. If today's AI systems are useful models for human learning, what they suggest is a picture that is associationist about the engine while remaining open about the surrounding machinery --- and agnostic about how much of the cognitive work is done by learning mechanisms shared across domains versus structures that are more domain-specific.

\bibliography{danbib}

\end{document}